# Soil Compaction Parameters Prediction Based on Automated Machine Learning Approach

Caner Erden[1*], Alparslan Serhat Demir[2], Abdullah Hulusi Kökçam[3], Talas Fikret Kurnaz[4], Ugur Dagdeviren[5]

[1] Corresponding author, Sakarya University of Applied Sciences, Faculty of Technology, Department of Computer Engineering, Sakarya, Türkiye, cerden@subu.edu.tr, ORCID: 0000-0002-7311-862X, cerden@subu.edu.tr

[2] Department of Industrial Engineering, Faculty of Engineering, Sakarya University, Sakarya, Türkiye, alparslanserhat@sakarya.edu.tr, ORCID: 0000-0003-3415-8116

[3] Department of Industrial Engineering, Faculty of Engineering, Sakarya University, Sakarya, Türkiye, akokcam@sakarya.edu.tr, ORCID: 0000-0002-4757-1594

[4] Technical Sciences Vocational School, Transportation Services, Mersin University, Mersin, Türkiye, fkurnaz@mersin.edu.tr, ORCID: 0000-0003-2079-8315

[5] Department of Civil Engineering, Faculty of Engineering, Kutahya Dumlupinar University, Kutahya, Türkiye, ugur.dagdeviren@dpu.edu.tr, ORCID: 0000-0002-4760-6574

**ABSTRACT**

Soil compaction is critical in construction engineering to ensure the stability of structures like road embankments and earth dams. Traditional methods for determining optimum moisture content (OMC) and maximum dry density (MDD) involve labor-intensive laboratory experiments, and empirical regression models have limited applicability and accuracy across diverse soil types. In recent years, artificial intelligence (AI) and machine learning (ML) techniques have emerged as alternatives for predicting these compaction parameters. However, ML models often struggle with prediction accuracy and generalizability, particularly with heterogeneous datasets representing various soil types. This study proposes an automated machine learning (AutoML) approach to predict OMC and MDD. AutoML automates algorithm selection and hyperparameter optimization, potentially improving accuracy and scalability. Through extensive experimentation, the study found that the Extreme Gradient Boosting (XGBoost) algorithm provided the best performance, achieving R-squared values of 80.4% for MDD and 89.1% for OMC on a separate dataset. These results demonstrate the effectiveness of AutoML in predicting compaction parameters across different soil types. The study also highlights the importance of heterogeneous datasets in improving the generalization and performance of ML models. Ultimately, this research contributes to more efficient and reliable construction practices by enhancing the prediction of soil compaction parameters.

## 1    INTRODUCTION

Since the soils are natural materials, their properties may vary for each construction site. In most construction sites that do not have the desired soil properties, improving the soil properties on-site is more appropriate due to technological and economic reasons. The compaction method is preferred, especially to improve the material properties of the soils used in road embankments and earth dam construction. Compaction is a stability method to increase unit volume weight by applying energy to the soil. With the application of the compaction method, the resistance of the soil against external influences increases while its permeability decreases.

The maximum dry density (MDD) ($\gamma_{d,max}$) and optimum moisture content (OMC) ($w_{opt}$), which play an important role in the compaction of engineering structures such as road embankments and earth dams, are determined on the compaction curve obtained as a result of Standard and Modified Proctor tests conducted in a laboratory. Obtaining the compaction curve requires proper sample preparation, compaction with a certain energy, measurement and calculation. Moreover, at least six tests are needed to determine the compaction curve accurately. Therefore, determining OMC and MDD values through laboratory experiments is time-consuming and laborious. To overcome this and determine the compaction parameters more effectively, alternative prediction models based on regression analyses using the soil index properties have been developed.

In parallel with the successful use of artificial intelligence (AI) techniques in many problems of geotechnical engineering, estimating compaction parameters using AI techniques has become a powerful alternative in recent years. Starting from the first studies, successful results have been achieved in estimating the compaction parameters by using various ML models such as deep neural network (DNN), extreme learning machine (ELM), support vector regression (SVR), genetic programming (GP), multi expression programming (MEP) and especially ANN-based ones [1], [2], [3], [4], [5], [6], [7], [8], [9], [10], [11], [12].

To our best knowledge, although AutoML approaches have been employed in geotechnical problems such as soil liquefaction [13], landslide susceptibility [14], [15], or slope stability classification [16], [17], no prior studies utilizing AutoML have been identified for the soil compaction prediction. To address these challenges and potentially achieve superior performance, this study presents an AutoML approach to predict OMC and MDD. We employ AutoML, a cutting-edge technique that automates the process of selecting and optimizing parameters of algorithms, for soil compaction prediction. AutoML has the potential to overcome limitations associated with manual hyperparameter tuning and explore a wider range of model architectures, potentially leading to improved accuracy and generalizability. Additionally, this study emphasizes that the ML model performance on estimating the compaction parameters largely depends on appropriate data selection and heterogeneous representation in terms of soil type.

## 2    DATA ACQUISITION AND ANALYSIS

The main dataset used in this study includes the data provided by Günaydın [4] for the training and testing purposes. The dataset consists of experimental results (compaction and index) of 126 soil samples. There are nine different soil classes in the dataset (SC, GM, GC, ML, MH, MI, CL, CH, CI). In the dataset of this study, liquid limit (LL), plastic limit (PL), gravel content (G%), sand content (S%), and fine content (F%) were used as input parameters, while OMC and MDD were used as output parameters. Upon examining the dataset, it was discovered that 11 data samples were duplicated. These data were removed from the dataset. Thus, analysis was carried out with the remaining 115 data. Among the 115 data, there are 62 fine-grained and 53 coarse-grained soil samples.

In addition, validation tests were carried out on different datasets in the literature, to both emphasize the importance of approaches that consider all influencing variables on the indirect estimation of compaction parameters and to demonstrate the robustness of the model proposed in the study. In this context, three different datasets were used. The first of these relatively heterogeneous in terms of soil type [18]. The second dataset consists predominantly fine-grained soils [19] while the third dataset consists of coarse-grained soils [20].

## 3    AUTOGLUON TABULAR PREDICTION

AutoGluon Tabular Prediction (ATP) is a Python library that uses ML techniques to predict target values in a column based on tables containing a dataset. It consists of two parts: TabularDataset and TabularPredictor. The first one is used to load the data, and the second one is used to train the models and obtain predictions. One of the reasons why AutoGluon is preferred over other libraries is that it corrects the definition errors made by other libraries in the data extraction processes. Additionally, it does not require intensive feature engineering regarding the features used in the model. It also simplifies the training process of the models by simplifying the hyperparameter setting [21].

In addition, AutoGluon simplifies ML model creation through presets and adjustable hyperparameters. Presets, pre-configured with optimized hyperparameter values, cater to specific goals like accuracy, speed, or memory usage, offering an accessible starting point, particularly for novice users. For instance, the "Best Quality" preset may enhance model performance by allocating more training time and conducting a broader hyperparameter search. Conversely, hyperparameters, individual settings within presets or models, allow for finer control and customization, influencing various aspects of the model's behavior and potentially enhancing performance through careful optimization. The decision between presets and hyperparameter adjustment depends on the user's experience level, desired control level, and project needs, such as handling unique data types or complex problems.

In this study, an AutoML approach is employed for predicting soil compaction parameters, leveraging numerous advantages offered by AutoML methodologies. AutoML techniques bring several benefits to the table, such as automating the model selection process, optimizing hyperparameters, and streamlining feature engineering tasks. These advantages contribute to enhanced efficiency and effectiveness in developing predictive models for soil compaction parameters. The automatic machine learning approach in this study is shown in Figure 1.

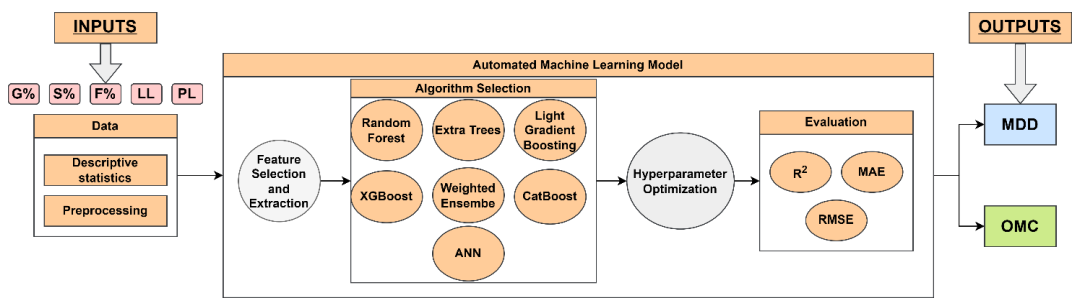

Figure 1. Automated Machine Learning (AutoML) Approach for Prediction of Soil Compaction Parameters

## 4    RESULTS AND DISCUSSION

In the domain of ML, selecting the most appropriate parameters for training a model is crucial. The automated ML could achieve this by choosing the effective configuration. Table 1 summarizes four different configurations we have defined for training a model using AutoGluon, allowing for a comparison of their potential strengths and weaknesses. First configuration uses the default settings offered by AutoGluon. It lacks any specified model name or preset, implying it might rely on the library's internal selection process for both the model architecture and hyperparameters. This is a good starting point for exploration, but it might not achieve the best possible performance compared to more tuned configurations. Second configuration leverages AutoGluon's built-in presets, specifically the "Best Quality" preset. This preset prioritizes achieving the highest possible accuracy on the validation set, potentially at the expense of training time or computational resources. This is a good option if maximizing accuracy is your primary concern but be aware of potential drawbacks like longer training times. Third configuration focuses on specifying the hyperparameter 'multimodal'. However, without knowing the specific model's name, it is difficult to pinpoint the exact impact of this hyperparameter. Some AutoGluon models might have built-in support for multimodal data, while others might require additional configuration. And the last configuration combines the elements from configurations 2 and 3. It utilizes the "Best Quality" preset for potentially higher accuracy and explicitly states the use of multimodal data through the 'multimodal' hyperparameter. This offers a balance between accuracy and tailoring the training process for multimodal data.

Table 1. Configurations used in AutoGluon

| Config No | Model Name | Preset | Hyperparameters |
|-----------|------------|--------|-----------------|
| 1 | Regular | - | |
| 2 | Presets | Best Quality | - |
| 3 | Hyperparameters | - | 'multimodal' |
| 4 | Both | Best Quality | 'multimodal' |

By employing these four configurations, a comparison can be made to evaluate their effectiveness in achieving the desired objectives, whether it be maximizing quality, handling multimodal data efficiently, or a combination of both. The evaluation might involve metrics such as accuracy, speed of training, resource utilization, etc. The best configuration depends on your specific goals and dataset characteristics.

### 4.1 Performance evaluation of trained models for OMC Prediction

Table 2 compares the R-squared performance of ML models for OMC prediction based on the 1st configuration. Notably, models like ExtraTreesMSE, LightGBMLarge, and RandomForestMSE exhibit higher test scores, indicating better performance on unseen data, though the validation scores help identify potential overfitting issues. WeightedEnsemble_L2 stands out with its high test and validation scores, suggesting strong overall performance, albeit at the cost of longer training times. Conversely, models such as KNeighborsDist demonstrate quick training times but may sacrifice performance.

Table 2. Performance of ML models for OMC prediction based on the 1st configuration

| Order | Model | Test Score | Validation Score | Training Time (in seconds) |
|-------|-------|------------|------------------|----------------------------|
| 1 | ExtraTreesMSE | 0.852 | 0.688 | 0.697 |
| 2 | LightGBMLarge | 0.851 | 0.637 | 1.332 |
| 3 | RandomForestMSE | 0.844 | 0.645 | 1.094 |
| 4 | XGBoost | 0.839 | 0.614 | 1.315 |
| 5 | WeightedEnsemble_L2 | 0.834 | 0.725 | 3.805 |
| 6 | NeuralNetTorch | 0.834 | 0.695 | 1.281 |
| 7 | CatBoost | 0.817 | 0.663 | 2.641 |
| 8 | NeuralNetFastAI | 0.796 | 0.714 | 1.840 |
| 9 | KNeighborsUnif | 0.783 | 0.472 | 3.176 |
| 10 | KNeighborsDist | 0.782 | 0.497 | 0.015 |
| 11 | LightGBMXT | 0.620 | 0.667 | 1.399 |
| 12 | LightGBM | 0.589 | 0.595 | 0.837 |

The R-squared performance comparison of OMC predictions for ML models in the 2nd configuration is provided in Table 3. Notably, RandomForestMSE_BAG_L1 achieves the highest test score of 0.892, followed closely by ExtraTreesMSE_BAG_L1 with a score of 0.873. However, LightGBMLarge_BAG_L1 exhibits the longest training time at 7.622 s, indicating higher computational cost. WeightedEnsemble_L2 demonstrates a strong validation score of 0.750, albeit with the longest training time of 32.887 s. Models like KNeighborsDist_BAG_L1 and KNeighborsUnif_BAG_L1 have the shortest training times but lower test and validation scores, suggesting potential performance trade-offs.

Table 3. Performance of ML models for OMC prediction based on the 2nd configuration

| Order | Model | Test Score | Validation Score | Training Time (in seconds) |
|---|---|---|---|---|
| 1 | RandomForestMSE_BAG_L1 | 0.892 | 0.687 | 0.690 |
| 2 | ExtraTreesMSE_BAG_L1 | 0.873 | 0.725 | 0.728 |
| 3 | LightGBMLarge_BAG_L1 | 0.859 | 0.666 | 7.622 |
| 4 | CatBoost_BAG_L1 | 0.848 | 0.717 | 5.442 |
| 5 | XGBoost_BAG_L1 | 0.846 | 0.703 | 5.000 |
| 6 | WeightedEnsemble_L2 | 0.839 | 0.750 | 32.887 |
| 7 | NeuralNetTorch_BAG_L1 | 0.828 | 0.706 | 11.187 |
| 8 | NeuralNetFastAI_BAG_L1 | 0.798 | 0.735 | 7.733 |
| 9 | KNeighborsDist_BAG_L1 | 0.757 | 0.615 | 0.005 |
| 10 | KNeighborsUnif_BAG_L1 | 0.747 | 0.566 | 0.010 |
| 11 | LightGBM_BAG_L1 | 0.687 | 0.605 | 5.239 |
| 12 | LightGBMXT_BAG_L1 | 0.684 | 0.628 | 5.216 |

Table 4 illustrates R-squared performance of ML models for OMC prediction based on the 3rd configuration. LightGBMLarge achieves the highest test score of 0.851, closely followed by XGBoost with a score of 0.839. NeuralNetTorch and CatBoost also perform well with test scores of 0.834 and 0.817, respectively. WeightedEnsemble_L2 demonstrates a competitive test score of 0.816, indicating robust performance. However, it comes with a relatively longer training time of 5.267 s. LightGBMXT, LightGBM, and other models exhibit lower test scores, suggesting comparatively weaker predictive capabilities.

Table 4. Performance of ML models for OMC prediction based on the 3rd configuration

| Order | Model | Test Score | Validation Score | Training Time (in seconds) |
|---|---|---|---|---|
| 1 | LightGBMLarge | 0.851 | 0.637 | 1.542 |
| 2 | XGBoost | 0.839 | 0.614 | 0.880 |
| 3 | NeuralNetTorch | 0.834 | 0.695 | 1.507 |
| 4 | CatBoost | 0.817 | 0.663 | 1.051 |
| 5 | WeightedEnsemble_L2 | 0.816 | 0.710 | 5.267 |
| 6 | LightGBMXT | 0.620 | 0.667 | 0.957 |
| 7 | LightGBM | 0.589 | 0.595 | 0.912 |

Table 5 presents the R-squared performance of ML models for OMC prediction based on the 4th configuration. XGBoost_BAG_L1_FULL achieves the highest test score of 0.891, followed closely by WeightedEnsemble_L2_FULL with a score of 0.888. CatBoost_BAG_L1_FULL also performs well with a test score of 0.867. NeuralNetTorch_BAG_L1_FULL, LightGBMLarge_BAG_L1_FULL, and other models exhibit lower test scores, suggesting comparatively weaker predictive capabilities. However, these models generally have shorter training times, indicating computational efficiency.

Table 5. Performance of ML models for OMC prediction based on the 4th configuration

| Order | Model | Test Score | Validation Score | Training Time (in seconds) |
|-------|-------|-----------|------------------|-----------------------------|
| 1 | XGBoost_BAG_L1_FULL | 0.891 | 0.739 | 0.053 |
| 2 | WeightedEnsemble_L2_FULL | 0.888 | 0.717 | 3.290 |
| 3 | CatBoost_BAG_L1_FULL | 0.867 | 0.706 | 0.230 |
| 4 | NeuralNetTorch_BAG_L1_FULL | 0.820 | 0.703 | 0.841 |
| 5 | LightGBMLarge_BAG_L1_FULL | 0.814 | 0.666 | 0.554 |
| 6 | LightGBM_BAG_L1_FULL | 0.765 | 0.628 | 0.618 |
| 7 | LightGBMXT_BAG_L1_FULL | 0.759 | 0.605 | 0.535 |

Table 6 provides insights into the importance of different features in influencing a model's output, along with statistical measures such as standard deviation, p-value, and confidence intervals. "LL" emerges as the most significant feature, demonstrating a high importance level with a low standard deviation and statistically significant p-value. "F%" and "S%" follow with moderate importance levels, albeit with slightly higher standard deviations and p-values. "PL" exhibits a lower importance level compared to others but still holds statistical significance. In contrast, "G%" displays negligible importance, even contributing negatively to the model output.

Table 6. Feature importance analysis for OMC model

| Index | Importance | Standard Deviation | P-value | Sample Size (n) | 99th Percentile High | 99th Percentile Low |
|-------|-----------|---------------------|---------|------------------|----------------------|---------------------|
| LL | 0.473 | 0.145 | 0.001 | 5 | 0.771 | 0.174 |
| F% | 0.139 | 0.081 | 0.009 | 5 | 0.306 | -0.029 |
| S% | 0.130 | 0.110 | 0.029 | 5 | 0.356 | -0.097 |
| PL | 0.069 | 0.035 | 0.006 | 5 | 0.141 | -0.002 |
| G% | -0.001 | 0.009 | 0.585 | 5 | 0.018 | -0.020 |

Comparison of actual and predicted values of both training and test sets for Günaydın (2009) OMC data and the prediction errors is provided in Figure 2.

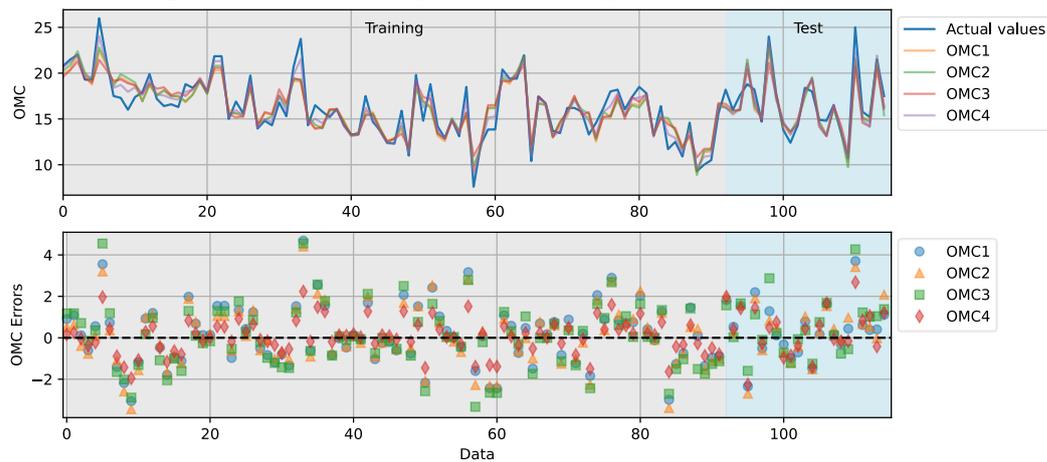

Figure 2. Comparison of actual and predicted values and errors of both training and test sets for Günaydın (2009) OMC data

## 4.2 Performance evaluation of trained models for MDD Prediction

Table 7 presents R-squared performance metrics of ML models for MDD prediction based on the 1st configuration. CatBoost emerges as the top-performing model with a test score of 0.781 and a validation score of 0.704, followed closely by WeightedEnsemble_L2 with slightly lower scores but a longer training time. ExtraTreesMSE and RandomForestMSE also perform reasonably well, with test scores above 0.74. However, XGBoost demonstrates comparatively lower performance with a test score of 0.711. NeuralNetFastAI and KNeighborsDist exhibit similar performance, while LightGBMLarge, KNeighborsUnif, and NeuralNetTorch perform moderately. LightGBMXT, LightGBM, and some other models show relatively lower test and validation scores, suggesting weaker predictive capabilities.

Table 7. Performance of ML models for MDD prediction based on the 1st configuration

| Order | Model | Test Score | Validation Score | Training Time (in seconds) |
|-------|-------|-----------|-----------------|----------------------------|
| 1 | CatBoost | 0.781 | 0.704 | 1.168 |
| 2 | WeightedEnsemble_L2 | 0.779 | 0.711 | 2.884 |
| 3 | ExtraTreesMSE | 0.743 | 0.665 | 0.667 |
| 4 | RandomForestMSE | 0.741 | 0.637 | 0.625 |
| 5 | XGBoost | 0.711 | 0.547 | 0.506 |
| 6 | NeuralNetFastAI | 0.705 | 0.624 | 0.938 |
| 7 | KNeighborsDist | 0.698 | 0.575 | 0.015 |
| 8 | LightGBMLarge | 0.674 | 0.508 | 1.648 |
| 9 | KNeighborsUnif | 0.672 | 0.558 | 0.015 |
| 10 | NeuralNetTorch | 0.652 | 0.621 | 0.872 |
| 11 | LightGBMXT | 0.566 | 0.557 | 0.773 |
| 12 | LightGBM | 0.551 | 0.543 | 0.649 |

Table 8 outlines R-squared performance metrics of ML models for MDD prediction based on the 2nd configuration. XGBoost_BAG_L1 emerges as the top-performing model with a test score of 0.804 and a validation score of 0.698, followed closely by WeightedEnsemble_L2 with slightly lower scores but a significantly longer training time. ExtraTreesMSE_BAG_L1 and CatBoost_BAG_L1 also demonstrate strong performance, with test scores above 0.791. However, KNeighborsDist_BAG_L1 exhibits notably lower test and validation scores, indicating weaker predictive capabilities compared to other models.

Table 8. Performance of ML models for MDD prediction based on the 2nd configuration

| Order | Model | Test Score | Validation Score | Training Time (in seconds) |
|-------|-------|-----------|-----------------|----------------------------|
| 1 | XGBoost_BAG_L1 | 0.804 | 0.698 | 5.505 |
| 2 | WeightedEnsemble_L2 | 0.798 | 0.737 | 26.984 |
| 3 | ExtraTreesMSE_BAG_L1 | 0.795 | 0.709 | 0.709 |
| 4 | CatBoost_BAG_L1 | 0.791 | 0.698 | 4.890 |
| 5 | NeuralNetFastAI_BAG_L1 | 0.771 | 0.724 | 7.838 |
| 6 | RandomForestMSE_BAG_L1 | 0.768 | 0.676 | 0.731 |
| 7 | NeuralNetTorch_BAG_L1 | 0.752 | 0.703 | 12.829 |
| 8 | KNeighborsDist_BAG_L1 | 0.713 | 0.575 | 0.000 |
| 9 | LightGBMLarge_BAG_L1 | 0.686 | 0.615 | 7.016 |
| 10 | KNeighborsUnif_BAG_L1 | 0.667 | 0.545 | 0.017 |
| 11 | LightGBM_BAG_L1 | 0.660 | 0.624 | 6.018 |
| 12 | LightGBMXT_BAG_L1 | 0.643 | 0.622 | 5.596 |

Table 9 presents R-squared performance metrics of ML models for MDD prediction based on the 3rd configuration. The models are assessed using test scores, validation scores, and training times. XGBoost emerges as the top-performing model with a test score of 0.785 and a validation score of 0.675, closely followed by WeightedEnsemble_L2 with slightly lower scores but a longer training time. CatBoost also exhibits strong performance, with test and validation scores close to those of XGBoost. LightGBMLarge, NeuralNetTorch, LightGBMXT, and LightGBM demonstrate relatively lower test and validation scores compared to the top-performing models, indicating weaker predictive capabilities in this configuration.

Table 9. Performance of ML models for MDD prediction based on the 3rd configuration

| Order | Model | Test Score | Validation Score | Training Time (in seconds) |
|-------|-------|------------|------------------|----------------------------|
| 1 | XGBoost | 0.785 | 0.675 | 0.836 |
| 2 | WeightedEnsemble_L2 | 0.782 | 0.707 | 3.844 |
| 3 | CatBoost | 0.781 | 0.704 | 2.197 |
| 4 | LightGBMLarge | 0.674 | 0.508 | 1.285 |
| 5 | NeuralNetTorch | 0.652 | 0.621 | 1.292 |
| 6 | LightGBMXT | 0.566 | 0.557 | 0.899 |
| 7 | LightGBM | 0.551 | 0.543 | 1.101 |

Similarly, Table 10**Hata! Yer işareti başvurusu geçersiz.** provides a feature importance analysis for the MDD prediction model. "LL" emerges as the most influential feature, with an importance value of 0.356 and a statistically significant p-value of 0.005. "PL" and "F%" also exhibit notable importance levels, with importance values of 0.162 and 0.148, respectively. However, "S%" and "G%" show comparatively lower importance levels.

Table 10. Feature importance analysis for MDD model

| Index | Importance | Standard Deviation | P-value | Sample Size (n) | 99th Percentile High | 99th Percentile Low |
|-------|-----------|--------------------|---------|-----------------|---------------------|---------------------|
| LL | 0.356 | 0.172 | 0.005 | 5 | 0.711 | 0.001 |
| PL | 0.162 | 0.062 | 0.002 | 5 | 0.289 | 0.035 |
| F% | 0.148 | 0.101 | 0.016 | 5 | 0.357 | -0.061 |
| S% | 0.056 | 0.064 | 0.060 | 5 | 0.188 | -0.075 |
| G% | 0.024 | 0.012 | 0.006 | 5 | 0.048 | -0.001 |

Comparison of actual and predicted values of both training and test sets for Günaydın (2009) MDD data and the prediction errors is provided in Figure 3.

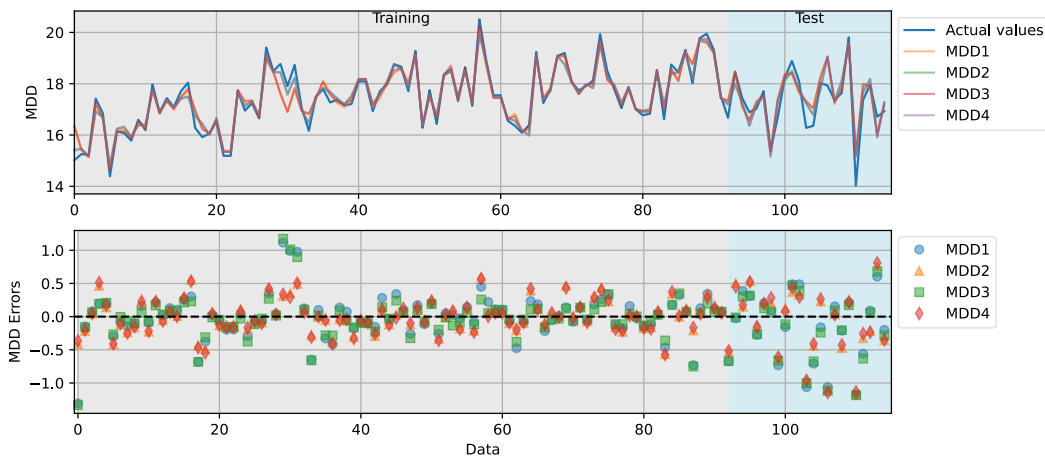

Figure 3. Comparison of actual and predicted values and errors of both training and test sets for Günaydın (2009) MDD data

## 5 CONCLUSION

This study evaluated the performance of a proposed AutoML approach for predicting soil compaction parameters, specifically MDD and OMC. Our experimental results identified XGBoost as the optimal algorithm, achieving a prediction accuracy of 0.804 for MDD and 0.891 for OMC on the test set. These results highlight the effectiveness and robustness of our approach across diverse soil types.

Beyond advancing the field of soil science, this research offers a clear pathway for integrating intelligent manufacturing and service systems into the construction industry. By automating the prediction of critical compaction parameters, our approach serves as a foundational component for intelligent construction workflows. For example, it could be integrated into smart infrastructure systems where sensors on compaction equipment collect real-time data.

This integration would enable real-time decision support, allowing automated adjustments to equipment settings to optimize compaction efforts and ensure quality control on the fly. The scalability of our AutoML model means it can be adapted for large-scale projects and continuously updated with new data to improve its predictive accuracy. In essence, this study transforms a traditional manual process into a data-driven, automated one, significantly improving the efficiency, reliability, and precision of construction practices.